\pdfoutput=1

\documentclass[11pt]{article}

\usepackage{emnlp2021}

\usepackage{times}
\usepackage{latexsym}
\usepackage{textcomp}

\usepackage[T1]{fontenc}

\usepackage[utf8]{inputenc}

\usepackage{microtype}

\usepackage{url}            %
\usepackage{booktabs}       %
\usepackage{amsfonts}       %
\usepackage{nicefrac}       %
\usepackage{microtype}      %

\usepackage{cuted}

\usepackage{enumitem}
\usepackage{bm}
\usepackage{rotating}
\usepackage{multirow}
\usepackage[percent]{overpic}

\usepackage{listings} %

\usepackage{amsmath,algorithm}
\usepackage[noend]{algpseudocode}

\usepackage{pifont}%
\newcommand{\cmark}{\ding{51}}%
\newcommand{\xmark}{\ding{55}}%

\definecolor{custom_red}{HTML}{EA6B66}
\definecolor{custom_green}{HTML}{97D077}

\title{SPARQLing Database Queries from Intermediate Question Decompositions}

\author{Irina Saparina \and Anton Osokin \\
  HSE University / Yandex / Moscow, Russia \\}

\renewcommand{\paragraph}{\textbf} %

\begin{document}
\maketitle
\begin{abstract}
To translate natural language questions into executable database queries, most approaches rely on a fully annotated training set.
Annotating a large dataset with queries is difficult as it requires query-language expertise.
We reduce this burden using grounded in databases intermediate question representations.
These representations are simpler to collect and were originally crowdsourced within the Break dataset \citep{wolfson-etal-2020-break}.
Our pipeline consists of two parts: a neural semantic parser that converts natural language questions into the intermediate representations and a non-trainable transpiler to the SPARQL query language (a standard language for accessing knowledge graphs and semantic web).
We chose SPARQL because its queries are structurally closer to our intermediate representations (compared to SQL).
We observe that the execution accuracy of queries constructed by our model on the challenging Spider dataset is comparable with the state-of-the-art text-to-SQL methods trained with annotated SQL queries.
Our code and data are publicly available.\footnote{\url{https://github.com/yandex-research/sparqling-queries}}\footnote{This version differs from the one presented at EMNLP-2021 in the figures but all conclusions still hold. The changes are caused by fixing several bugs in the decoding process, usage of the GraPPa tokenization (affect our model) and SQL-SQL comparison (affect the baselines).}
\end{abstract}

\section{Introduction}
The difficulty of collecting and annotating datasets for the task of translating a natural language question to an executable database query is a significant obstacle to the progress of the technology.
The most popular multi-database text-to-SQL dataset, Spider \citep{yu-etal-2018-spider}, has 10K questions, which is smaller compared to question answering datasets of other types: %
the DROP dataset with text paragraphs has 97K questions~\citep{dua-etal-2019-drop} and the GQA dataset with images has 22M questions~\citep{hudson2019gqa}.
The Spider dataset was created by 11 Yale students proficient in SQL, and it is difficult to scale such a process up. %

Recently, \citet{wolfson-etal-2020-break} proposed the Question Decomposition Meaning Representation, QDMR, which is a way to decompose a question into a list of ``atomic'' steps representing an algorithm for answering the question.
Importantly, they developed a crowdsourcing pipeline to annotate QDMRs and showed that it can be used at scale: they
collected 83K QDMRs for questions (all in English) coming from different datasets (including Spider) and released them in the Break dataset.

QDMRs resemble database queries but are not connected to any execution engine and cannot be run directly.
Moreover, QDMRs were collected when looking only at questions and thus have no information about the database structure.
Entities mentioned in QDMR steps usually have counterparts in the corresponding database but do not have links to them (grounding). %

\begin{figure*}[t!]
\centering
\includegraphics[width=\textwidth,clip=true,trim=0mm 1mm 18mm 2mm]{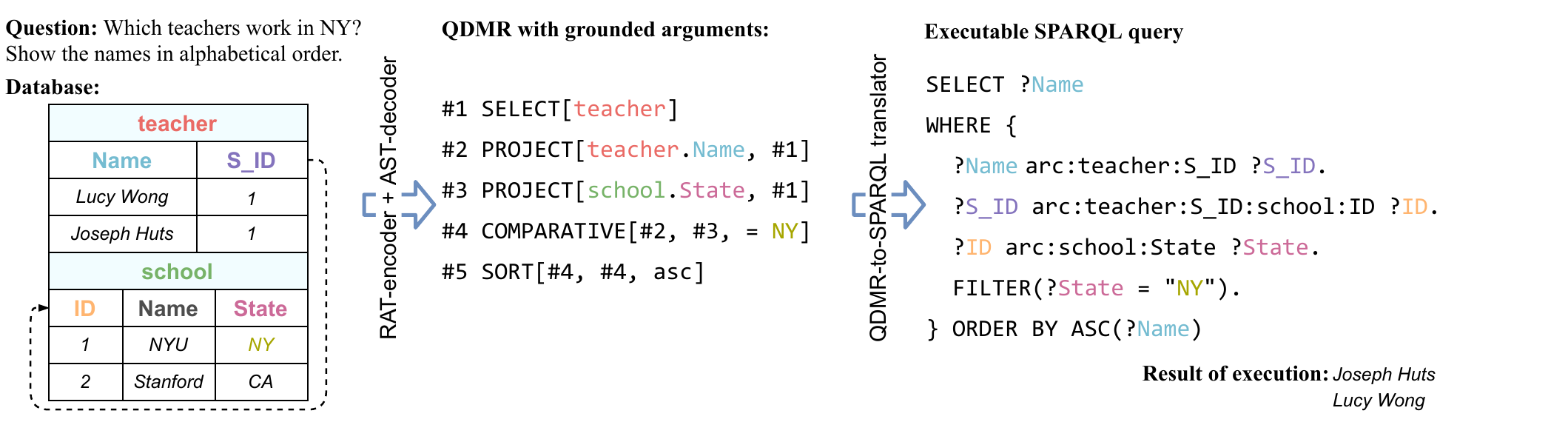}
\caption{Overall map of our approach: we feed a question and a database schema into the encoder-decoder model to obtain the grounded QDMR.
The grounded QDMR is then fed into our QDMR-to-SPARQL translator to obtain an executable SPARQL query.
The generated query is executed on the database in the RDF format.
\label{fig:schema}}
\end{figure*}

In this paper, we build a system for translating a natural language question first into QDMR and then into an executable query.
We use modified QDMRs, where the entities described with text are replaced with their database groundings.
Our system consists of two translators: a neural network for text-to-QDMR and a non-trainable QDMR-to-SPARQL transpiler.
See Figure~\ref{fig:schema}, for an illustration of our system.

In the text-to-QDMR part, we use an encoder-decoder model.
Our encoder is inspired by RAT-transformer \citep{wang-etal-2020-rat} and uses BERT \citep{devlin-etal-2019-BERT} or GraPPa \citep{yu-etal-2021-GraPPa}.
Our decoder is a syntax-guided network~\citep{yin-neubig-2017-syntactic} designed for our version of the QDMR grammar. %
We trained this model with full supervision, for which we automatically grounded QDMRs for a subset of Spider questions.

\begin{figure}[t!]

\includegraphics[width=\columnwidth,clip=true,trim=0mm 7mm 0mm 0mm]{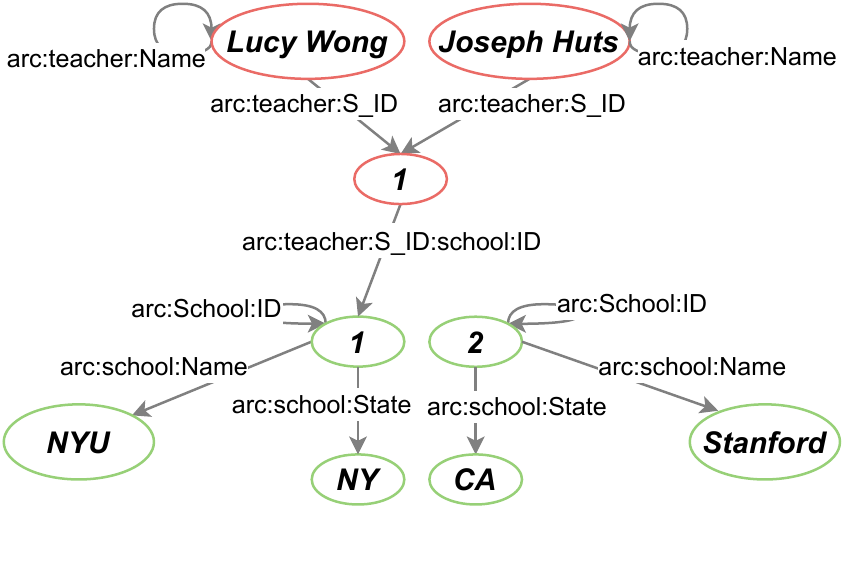}
\caption{Database from Figure~\ref{fig:schema} converted to the RDF format (the RDF graph).
The \textcolor{custom_red}{red} nodes correspond to the values from \textbf{\textcolor{custom_red}{teacher}} table, the \textcolor{custom_green}{green} ones - to the values from \textbf{\textcolor{custom_green}{school}} table.
Arcs correspond to the relations between primary key and other values of the same row (\texttt{arc:tbl:col}) and along the foreign keys (\texttt{arc:t\_src:c\_src:t\_tgt:c\_tgt}).
\label{fig:rdf-graph}}
\end{figure}

In the second part of the system, our goal was to translate grounded QDMRs into one of the existing query languages to benefit from the efficiency of database software.
The most natural choice would be to use SQL, but designing such a translator was difficult due to structural differences between QDMR and SQL.
Instead, we implement a translator from QDMR to SPARQL,\footnote{SPARQL is a recursive acronym for SPARQL Protocol and RDF Query Language.} which is a query language for databases in the Resource Description Framework (RDF) format \citep{sparql08,sparql}.
SPARQL is a standard made by the World Wide Web Consortium and is recognized as one of the key technologies of the semantic web.
See Figure~\ref{fig:rdf-graph} for an example of the RDF database.

We evaluated our system with the execution accuracy metric on the Spider dataset (splits by~\citealp{wolfson-etal-2020-break}) and compared it with two strong baselines: text-to-SQL systems BRIDGE \citep{lin-etal-2020-bridging} and SmBoP \citep{rubin2020smbop} from the top of the Spider leaderboard.
On the cleaned-up validation set, our system outperforms both baselines.
On the test set with original annotation, our system is in-between the baselines.
Additionally, we experimented with training our models on extra data: items from Break without schema but with QDMRs.
This teaser experiment showed potential for further improvements.

This paper is organized as follows.
Sections~\ref{sec:qdmr2sparql_main} and~\ref{sec:text2qdmr} present two main parts of our system.
Section~\ref{sec:experiments} contains the experimental setup, Section~\ref{sec:results}~-- our results.
We review related works in Section~\ref{sec:related_works} and conclude in Section~\ref{sec:conclusion}.

\section{QDMR-to-SPARQL translator} \label{sec:qdmr2sparql_main}
\subsection{QDMR logical forms}
Question Decomposition Meaning Representation (QDMR) introduced by \citet{wolfson-etal-2020-break} is an intermediate format between a question in a natural language (tested in English) and an executable query in some formal query language. %
QDMR is a sequence of steps, and each step corresponds to  a separate logical unit of the question (see Table~\ref{tab:qdmr_examples}).
A QDMR step can refer to one of the previous steps, allowing one to organize the steps into a graph. %

We work with QDMR logical forms (LF), which can be automatically obtained from the text-based QMDRs, e.g., with the rule-based method of~\citet{wolfson-etal-2020-break}.
Steps of a logical form are derived from the corresponding steps of QDMR.
Each step of LF includes an operator and its arguments.
We show some operators in Table~\ref{tab:qdmr_ops} and provide the full list in Appendix~\ref{sec:qdmr2sparql}.\footnote{Differently from \citet{wolfson-etal-2020-break} we merged the operation \texttt{FILTER} into \texttt{COMPARATIVE} due to their similarity and excluded \texttt{ARITHMETIC}, \texttt{BOOLEAN} and undocumented \texttt{COMPARISON} because they are extremely rare in the Spider part of Break.}
\begin{table}[t!]
    \centering
    \begin{tabular}{@{}l@{\;\;}l@{}}
        \toprule %
        Question: &  \begin{minipage}[c]{55mm}
        \footnotesize For each state, how many teachers are there? %
        \end{minipage}
        \\
        \midrule
        \begin{minipage}{20mm} 
        QDMR \\
        (Break)
        \end{minipage}
        &
        \begin{minipage}{55mm} 
        \footnotesize
        \texttt{\#1} return states\\
        \texttt{\#2} return teachers in \texttt{\#1}\\
        \texttt{\#3} return number of \texttt{\#2} for each  \texttt{\#1}\\
        \texttt{\#4} return \texttt{\#1} and \texttt{\#3}
        \end{minipage}
        \\
        \midrule
        \begin{minipage}{20mm} 
        QDMR \\ logical form\\ (Break)
        \end{minipage}
        & 
        \begin{minipage}{55mm} 
        \footnotesize
        \texttt{\#1 SELECT[}states\texttt{]}\\
        \texttt{\#2 PROJECT[}teachers in \texttt{\#REF, \#1]}\\
        \texttt{\#3 GROUP[count, \#2, \#1]}\\
        \texttt{\#4 UNION[\#1, \#3]}
        \end{minipage}
        \\
        \midrule
        \begin{minipage}{20mm} 
        grounded \\ QDMR\\
        (ours)
        \end{minipage}
        & 
        \begin{minipage}{55mm} 
        \footnotesize
        \texttt{\#1 SELECT[School.State]}\\
        \texttt{\#2 PROJECT[teacher, \#1]}\\
        \texttt{\#3 GROUP[count, \#2, \#1]}\\
        \texttt{\#4 UNION[\#1, \#3]}
        \end{minipage}
        \\
        \bottomrule
    \end{tabular}
    \caption{Examples of different QDMR formats: textual QDMR, QDMR logical form (from Break) and our version of QDMR with grounded arguments. \label{tab:qdmr_examples}}
\end{table}

\subsection{Grounding QDMRs in databases}
QDMR logical forms are similar to the programmed queries but are not connected to any execution engine and cannot be executed directly.
To execute these LFs using knowledge from a database, one needs to associate their arguments with the entities of the database: tables, columns, values.
We refer to this association as grounding and provide the details below.

Arguments of LF operators can be of different types (see Table~\ref{tab:qdmr_ops} and Appendix~\ref{sec:qdmr2sparql}) and some types require groundings.
Type~\texttt{ref} indicates a reference to one of the existing LF steps.
Type~\texttt{text} corresponds to a text argument that needs to be grounded to a table, column or value in the database.
Type~\texttt{choice} corresponds to the choice among a closed list of possible options, and type~\texttt{bool} corresponds to the True/False choice. %

There are also a few edge cases that require special processing.
First, the \texttt{value} argument of the \texttt{COMPARATIVE} operator can be either \texttt{ref} or \texttt{text}. %
Second, the \texttt{operator} argument of \texttt{AGGREGATE}/\texttt{GROUP} can actually be grounded to a column.
We introduced this exception because a database can contain only the aggregated information without information about individual instances.
As the QDMR annotation is built without looking at the database it cannot distinguish the two cases. In the example of Table~\ref{tab:qdmr_examples}, if the database has a column \texttt{num\_teachers} in the table~\texttt{school} we would need to ground \texttt{count} to the column  \texttt{num\_teachers}.

We describe our procedure for annotating LF arguments with groundings in Section~\ref{sec:annotate_groundings}.

\subsection{Executable queries in SPARQL}
To convert a QDMR LF with grounded step arguments into an actually executable query, it is beneficial to translate QDMR into one of the existing query languages to use an existing efficient implementation at the test time. 
In this paper, we translate QDMR queries into SPARQL, a language for querying databases in the graph-based RDF format \citep{sparql08,sparql}.
Next, we briefly overview the RDF database format and SPARQL and then describe our algorithm for translating grounded LFs into SPARQL queries.

\textbf{RDF format.}
In RDF, data is stored as a directed multi-graph, where the nodes correspond to the data elements and the arcs correspond to relations.
RDF-graphs are usually defined by sets of subject-predicate-object triples, where each triple defines an arc: the subject is the source node, the predicate is the type of relation and the object is the target node.

\textbf{Relational data to RDF.}
To evaluate our approach on the Spider dataset containing relational databases (in the SQLite format), we convert relational databases to the RDF format.
The conversion is inspired by the specification of~\citet{arenas2012rdb2rdf}.
For each table row of the relational database, we add to the RDF graph a set of triples corresponding to each column.
For the primary key column\footnote{For simplicity, we assume that each table has a single-column primary key (otherwise, we add a new \texttt{ID} column).
} \texttt{key} of a table \texttt{tbl}, we create a triple with the self-link \texttt{arc:tbl:key} pointing from the key element to itself.
For any other column \texttt{col} in the table \texttt{tbl}, we create a triple with the separate edge type \texttt{arc:tbl:col}, which connects the primary key element of a row to the corresponding element in \texttt{col}.
For each foreign key of the database, we create an arc type
\texttt{arc:t\_src:c\_src:t\_tgt:c\_tgt}
(here the target column \texttt{c\_tgt} has to be a key). %
Then we add to the RDF graph the triples with these foreign-key relations.
See Figure~\ref{fig:rdf-graph} with an example of the RDF graph for the database of Figure~\ref{fig:schema}.

\begin{table*}[t!]
    \centering
    \footnotesize
    \begin{tabular}{@{}l@{\;\;}l@{\;\;}l@{}}
        \toprule
         \textbf{Operator} & \textbf{Arguments:their types} & \textbf{Description} \\
        \midrule
         \multirow{1}{*}{\texttt{SELECT}} & [\texttt{subj}:\texttt{text}, \texttt{distinct}:\texttt{bool}] & Select \texttt{subj} (possibly \texttt{distinct} values) \\
        \midrule
         \multirow{1}{*}{\texttt{PROJECT}} & [\texttt{proj}:\texttt{text}, \texttt{subj}:\texttt{ref}] & Select \texttt{proj} related to \texttt{subj}  (possibly \texttt{distinct} values)\\
         \midrule
         \multirow{2}{*}{\texttt{COMPARATIVE}} & [\texttt{subj}:\texttt{ref},  \texttt{attr}:\texttt{ref},  \texttt{comp}:\texttt{choice}, & Select \texttt{subj} such that related \texttt{attr} \texttt{comp}ares (using $=$, $\neq$, \\
         & \texttt{value}:\texttt{text}/\texttt{ref}, \texttt{distinct}:\texttt{bool}] & $>$, $<$, $\geq$, $\leq$,  \texttt{like}) to \texttt{value} (possibly \texttt{distinct} values) \\
         \midrule
         \multirow{1}{*}{\texttt{GROUP}} & [\texttt{subj}:\texttt{ref}, \texttt{attr}:\texttt{ref}, \texttt{op}:\texttt{choice}]  & Group \texttt{subj} such that \texttt{attr} has same values (aggr. with \texttt{op}) \\
         \bottomrule
    \end{tabular}
    \caption{QDMR operators and their arguments with types. See Appendix~\ref{sec:qdmr2sparql} for the full version~--Table~\ref{tab:qdmr_ops_full}.}
    \label{tab:qdmr_ops}
\end{table*}

\textbf{SPARQL.} In a nutshell, a SPARQL query is a set of triple patterns where some elements are replaced with variables.
The execution happens by searching the RDF graph for subgraphs that match the patterns.
For example, a query 
\begin{lstlisting}[columns=fullflexible,keepspaces=true,basicstyle=\ttfamily,aboveskip=0mm,belowskip=0mm]
SELECT ?State WHERE {
  ?ID arc:school:State ?State.}
\end{lstlisting}
to the RDF graph of Figure~\ref{fig:rdf-graph} searches for pairs of nodes that are connected with arcs of type \texttt{arc:school:State}.
Entries starting with symbol \texttt{?} represent variables.
See Figure~\ref{fig:schema} for an example of a more complicated query.

SPARQL also supports subqueries and aggregators, the \texttt{GROUP}, \texttt{SORT}, \texttt{UNION}, \texttt{MINUS} keywords, etc.
See, e.g., the Wikidata SPARQL tutorial\footnote{\url{https://www.wikidata.org/wiki/Wikidata:SPARQL_tutorial}} for a detailed overview of SPARQL features.

\textbf{Translating grounded QDMR to SPARQL.}
We implemented a translator from a grounded QDMR LF into SPARQL.
Note that LFs do not have a formal specification defining the execution, so
our translator fills in the formal meaning. %
Our translator recursively constructs graph patterns that contain a result of LF steps.
When processing a step, the method first constructs one or several patterns for the step arguments and then connects them into another pattern.
At the beginning of the process, we request the method to construct the pattern containing the last QDMR step, which corresponds to the query output.
We provide the details of our translator in Appendix~\ref{sec:qdmr2sparql}.

\section{Text-to-QDMR parser} \label{sec:text2qdmr}
In this section, we describe our approach to generating a grounded QDMR LF from a given question and a database schema.
Our encoder consists of BERT-like pretrained embeddings \citep{devlin-etal-2019-BERT,yu-etal-2021-GraPPa} and a relation-aware transformer~\citep{wang-etal-2020-rat}.
Our decoder is an LSTM model that generates an abstract syntax tree in the depth-first traversal order \citep{yin-neubig-2017-syntactic}.

\subsection{Encoder}
In our task, the input is a sequence of question tokens and a set of database entities eligible for grounding: tables, columns, and  extracted values.

To choose values from a database, we use string matching between question tokens and database values (see Appendix \ref{sec:implementation}).  Additionally, we extract numbers and dates from the question that can be valid comparative values not from the database.
To avoid ambiguity of the encoding, we combine the multiple identical values from different columns into one. %

Following \citet{huang-etal-2018-natural, zhang-etal-2019-editing, wang-etal-2020-rat}, the input tokens of four types (question, table, column and value) are interleaved with \texttt{[SEP]}, combined into a sequence and encoded: we experiment with BERT \citep{devlin-etal-2019-BERT} and GraPPa \citep{yu-etal-2021-GraPPa}.
The obtained representations are fed into the relation-aware transformer, RAT \citep{wang-etal-2020-rat}.

\paragraph{RAT module.} RAT \citep{wang-etal-2020-rat} is based on relation-aware self-attention layer \citep{shaw-etal-2018-self} .
Unlike the standard self-attention in the transformer model \citep{Vaswani2017}, this layer explicitly adds embeddings $\boldsymbol{r}_{i j}$ 
that encode relations between two inputs $\boldsymbol{x}_{i}, \boldsymbol{x}_{j}$.
The RAT self-attention weights are $\alpha_{i j}=\text{softmax} \Big( \frac{\boldsymbol{x}_{i} W_{Q}\left(\boldsymbol{x}_{j} W_{K}+\boldsymbol{r}_{i j}\right)^{\top}}{\sqrt{d}} \Big)$,
where $W_{K}$, $W_{Q}$, $d$ are the standard self-attention parameters. 

The relations between the columns and tables come from the schema structure, e.g., the table~-- primary key and foreign key relations.
We also have relations based on matches: question~-- table and question~-- column matches based on the n-gram comparison  
\citep{guo-etal-2019-towards} and question~-- value matches from our value extracting procedure.

\subsection{Decoder}
The decoder is a recurrent model with LSTM cells that generates an abstract syntax tree (AST) in the depth-first traversal order \citep{yin-neubig-2017-syntactic}.
At each prediction, the decoder selects one of the allowed outputs, the list of allowed outputs is defined by our QDMR grammar (see Appendix \ref{sec:grammar}). The output can be the grammar rule (transition to a new node in AST), the grounding choice or the previous step number (leaf nodes in AST).

To predict grammar rules, we use the same modules as in the RAT-SQL model \citep{wang-etal-2020-rat}.
The decoder predicts comparator, aggregator and sort directions using the output of MLP.
For table, column or value grounding, we use the pointer network attention mechanism \citep{vinyals2015pointer}.
To predict a reference to a previous QDMR step, we use an MLP with a mask in the output softmax.
To avoid incorrect QDMR output, we use several constraints in the decoding process. Most of them are in the prediction of comparative arguments, e.g., we check type consistency (see Appendix \ref{sec:decoding_rules}).

\subsection{Training} 
We follow the RAT-SQL \citep{wang-etal-2020-rat} training procedure in the main aspects. 
We use the standard teacher forcing approach for autoregressive models.
We found that an additional alignment loss proposed for RAT-SQL did not lead to any improvements in our case, so we trained the models with the cross-entropy loss with label smoothing.
See Appendix~\ref{sec:implementation} for implementation details.

\paragraph{Augmentations.} We randomly permute tables, columns and values when training. 
We experimented with a random choice of QDMR graph linearization at training but did not observe performance improvements.
We also tried to randomly select one of the multiple available QDMR groundings, but it did not help as well.

\section{Experiment setup}
\label{sec:experiments}
\subsection{Data}
For training and evaluation, we use the part of the Break dataset that corresponds to the Spider dataset.\footnote{The Break dataset also contains QDMRs for other text-to-SQL datasets, e.g., single-database ATIS and GeoQuery.
Comparison in the regime of fine-tuning on a specific database is also interesting, but baseline and our codebases failed due to the limitations of the SQL parsers (coming from Spider).
This issue might be resolved by switching to a different SQL parser but it appeared technically infeasible at the time of writing.} %
Data includes questions and databases from Spider, QDMR logical forms from Break and groundings that we collected.
Automatic grounding annotation is challenging, but we are able to annotate with target groundings more than 60\% of the Break data (see Section~\ref{sec:annotate_groundings}). 
Our splits are based on the Break splits but take into account the grounding annotation.
The Break dataset does not include the Spider test, as it is hidden, while the Break dev and test are the halves of the Spider dev.
The gold QDMR and grounding annotation on the Break test is also hidden.
The overall dataset statistics are shown in Table~\ref{tab:dataset}.

\begin{table}[t]
\centering
\normalsize
\begin{tabular}{@{}lccc@{}}
\toprule %
\textbf{Dataset} & \textbf{Train} & \textbf{Dev} & \textbf{Test}\\
\midrule %
Original Spider & 8695 & 1034 & 2147 \\
- with Break & 6921 & 502 & 521  \\
- with groundings & 4350 & 445 & - \\
\bottomrule %
\end{tabular}
\caption{Dataset statistics for the original Spider, part of Spider with QDMR annotations from Break and part of Spider with QDMRs and groundings.
Break dev and test are splits of original Spider dev.
Break test is hidden, so we do not have annotation for this part.}
\label{tab:dataset}
\end{table}

We fixed typos and annotation errors in some train and dev examples.
We also corrected some databases on train and dev: we deleted trailing white spaces in values (led to mismatches between SQL query and database) and added missing foreign keys (necessary for our SPARQL generator) based on the procedure of \citet{lin-etal-2020-bridging}.
We kept the test questions and SQL queries unchanged from the original Spider dataset, which implied that some dataset errors could degrade comparisons of SQL and SPARQL results. 

\subsection{Annotating Groundings for LFs\label{sec:annotate_groundings}}

We process LFs from the Break dataset in several stages.
At the first stage, we iterate over all the operators and make their arguments compatible with our specification (see  Table~\ref{tab:qdmr_ops}).

At the second stage, we collect candidate groundings for each argument that requires grounding.
At this stage, we use all available sources of information: text-based similarity between the text argument and the names of the database entities, the corresponding SQL query from Spider, explicit linking between the question tokens and the elements of the schema released by~\citet{lei-etal-2020-examining}.
Importantly, we can match the output of LF to the output of the SQL query and propagate groundings inside LF, which allows to obtain many high-quality groundings.
At the third stage, we use collected candidate groundings and group them in all possible ways to obtain candidate LFs with all arguments grounded.
Then, for each candidate LF, we run our QDMR-to-SPARQL translator and execute the obtained query.
We accept the candidate if there are no failures in the pipeline and the result of the SPARQL query equals the result of the SQL one. 
Finally, we included the question in the dataset if we had accepted at least one grounded LF. %
Note that we can accept several versions of grounding for each question.
We cannot figure out which one is better at this point, so we can either pick one randomly or use all of them at training.

\subsection{Evaluation Metric}
For evaluation on the Spider dataset, most text-to-SQL methods use the metric called exact set matching without values.
This metric compares only some parts of SQL queries, e.g., values in conditions are not used, and sometimes incomplete non-executable queries can achieve high metric values.
As our approach does not produce any SQL query at all, this metric is not applicable.

Instead, we use an execution-based metric, a.k.a.\ execution accuracy.
This metric compares the results produced by the execution of queries (allowing arbitrary permutation on the output columns).
Recently, the Spider leaderboard  started supporting this metric, but submitting directly to the leaderboard is still not possible for us because the exposed interface requires SQL queries.
We modify the Spider execution accuracy evaluation in such a way that it can support any query language that can be executed and provide results.
When comparing the results of SPARQL to the results of SQL, we faced several challenges:
\begin{itemize}[noitemsep,topsep=0mm,leftmargin=*]
    \item the order of output columns in SQL does not match the order in the question;
    \item in Spider, when selecting relations w.r.t. argmin or argmax there is no consistent policy whether to pick all the rows satisfying the constraints or only one of them;
    \item the order of rows in the output of SQL is stable, but the order of rows in the output of SPARQL varies depending on minor launching conditions;
    \item in SPARQL, sorting is unstable (can arbitrarily change elements with equal sorting key values), but SQL sorting is stable; %
\end{itemize}
The first two points can make SQL-to-SQL comparisons invalid as well, and the others affect only SQL-to-SPARQL comparisons.

To resolve these issues, we implemented the metric supporting SQL-to-SQL, SQL-to-SPARQL, SPARQL-to-SPARQL comparisons with the following properties:
\begin{itemize}[noitemsep,topsep=0mm,leftmargin=*]
\item %
we reorder the columns of the outputs based on the columns the output values come from. If the matching fails, we try to compare output tables with the given order of columns; %
\item if one of the outputs is from an SQL query ending with ``ORDER BY$\cdot\cdot\cdot$LIMIT 1'', we check that the produced one row is contained in another output; %
\item if one of the outputs has done unstable sorting, we allow it to provide a key w.r.t.\ which the sorting was done and try to match the order of the rows in another output by swapping the rows with identical sorting-key values; %
\item before comparison, we extract the column types from both outputs and convert each value to the standardized representation.
\end{itemize}

\begin{table}[t]
\centering
\normalsize
\begin{tabular}{@{}lcccc@{}}
\toprule %
\textbf{Model} & \textbf{Train} & \textbf{Pretrain} & \textbf{Dev} & \textbf{Test}\\
\midrule %
BRIDGE & full & BERT & 71.5 & 64.5 \\
SmBoP & full & GraPPa & 78.2 & \textbf{66.4} \\
\midrule %
BRIDGE & subset   & BERT &   71.7  &  62.2 \\
SmBoP & subset  & GraPPa & 76.4 & \textbf{66.4} \\
\midrule %
Ours & subset  & BERT & 81.1 & 60.1 \\
Ours & subset  & GraPPa  & \textbf{82.0}  & 62.4\\ 
\bottomrule %
\end{tabular}
\caption{Execution accuracy of our model compared to state-of-the-art text-to-SQL methods on our development and test sets.}
\label{tab:sparql-sql}
\end{table}

\begin{table}[t]
\centering
\normalsize
\begin{tabular}{@{}lcccc@{}}
\toprule %
 \textbf{Train} &\textbf{Pretrain} & \textbf{Augs} & \textbf{Dev} & \textbf{Test}\\
\midrule %
subset & BERT &  +  & 81.1 & 60.1\\
full Break & BERT &  - & 76.4 & 63.3  \\
full Break & BERT &  + & 78.9 & 64.9 \\
\midrule
subset &  GraPPa & +  & \textbf{82.0}  & 62.4 \\
full Break & GraPPa & - & 81.3 & 64.3 \\
full Break & GraPPa &  +  & 81.6 & \textbf{65.3} \\
\bottomrule %
\end{tabular}
\caption{Execution accuracy of our model trained on only the Spider subset of Break compared to using additional data from Break (on our development and test sets).}
\label{tab:full_break}
\end{table}

\section{Results}
\label{sec:results}
\subsection{Comparison with text-to-SQL methods}
First, we compare our approach to state-of-the-art text-to-SQL methods (that generate full executable queries) BRIDGE \citep{lin-etal-2020-bridging} and SmBoP \citep{rubin2020smbop}, both from the top of the Spider leaderboard.
See Table \ref{tab:sparql-sql} for the results.
As our training data includes only 50\% of the original Spider train, we add to the comparison BRIDGE and SmBoP models trained on the same data subset.
We use the official implementations of both models. %
All models are trained together with finetuning pretrained contextualized representations: BRIDGE encoder uses BERT, SmBoP encoder uses GraPPa, our model has both BERT and GraPPa versions.

We choose the final model of each training run of our system based on the best dev result from the last 10 checkpoints with the step of 1000 iterations.
For BRIDGE and SmBoP, we used the procedures provided in the official implementations (they similarly look at the same dev set).
The estimated std of our model is $0.9$ on the dev set (estimated via retraining our BERT-based model with 5 different random seeds).

On the development set, our models achieve better execution accuracy than text-to-SQL parsers even trained on full Spider data.
On the test set, our models outperform BRIDGE but not SmBoP when trained on the same amount.
See Table~\ref{tab:qdmr_results} for qualitative results of our GraPPa-based model.

We did not include the results of RAT-SQL \citep{wang-etal-2020-rat} in Table~\ref{tab:sparql-sql}, because this model was trained to optimize exact set matching without values, so the model output contains placeholders instead of values.
The model trained on full Spider reproduces the exact matching scores shown by \citet{wang-etal-2020-rat} but gives only $40.2\%$ execution accuracy on dev and $39.9\%$ on test.
Correct predictions mostly came from correct SQL queries without values.
We also tried the available feature of value prediction in the official implementation of RAT-SQL and obtained better execution accuracy scores ($48.5\%$  on dev and $46.4\%$ on test), but they were still very low.

\begin{table}[p]
    \centering
    \begin{tabular}{@{}l@{\;\;}l@{}}
        \toprule %
        Q: &  \begin{minipage}[c]{65mm}
        \footnotesize 
        How many concerts are there in year 2014 or 2015? %
        \end{minipage}\\
      \midrule      
        SQL: & 
        \begin{minipage}{65mm} 
        \footnotesize \texttt{SELECT count(*) FROM concert} \\
        \texttt{WHERE YEAR = 2014 OR YEAR = 2015}
        \end{minipage}
        \\
        \midrule
        \begin{tabular}{@{}l@{\;\;}}Ours \\ \cmark \end{tabular}
        & 
        \begin{minipage}{65mm} 
        \footnotesize
        \texttt{\#1 SELECT[concert]}\\
        \texttt{\#2 PROJECT[concert.Year, \#1]}\\
        \texttt{\#3 COMPARATIVE[\#1,\#2,=2014]}\\
        \texttt{\#4 COMPARATIVE[\#1,\#2,=2015]} \\ %
        \texttt{\#5 UNION[\#3, \#4]}\\
        \texttt{\#6 AGGREGATE[count, \#5]}
        \end{minipage}
        \\
        \midrule
        Q: &  \begin{minipage}{65mm}
        \footnotesize Show location and name for all stadiums with a capacity between 5000 and 10000. %
        \end{minipage}\\
       \midrule      
        SQL: & 
        \begin{minipage}{65mm} 
        \footnotesize \texttt{SELECT location, name FROM stadium} \\
        \texttt{WHERE capacity}\\
        \texttt{BETWEEN 5000 AND 10000}
        \end{minipage}
        \\
        \midrule
        \begin{tabular}{@{}l@{\;\;}}Ours \\ \cmark \end{tabular} &
        \begin{minipage}{65mm} 
        \footnotesize
        \texttt{\#1 SELECT[stadium]}\\
        \texttt{\#2 PROJECT[stadium.Capacity, \#1]}\\
        \texttt{\#3 COMPARATIVE[\#1,\#2, $\geq$5000]}\\
        \texttt{\#4 COMPARATIVE[\#1,\#2, $\leq$10000]} \\ %
        \texttt{\#5 INTERSECTION[\#1, \#3, \#4]}\\
        \texttt{\#6 PROJECT[stadium.Location, \#5]}\\
        \texttt{\#7 PROJECT[stadium.Name, \#5]}\\
        \texttt{\#8 UNION[\#6, \#7]}
        \end{minipage}
        \\
        \midrule
        Q: &  \begin{minipage}{65mm}
        \footnotesize What is the year that had the most concerts? %
        \end{minipage}\\
      \midrule      
        SQL: & 
        \begin{minipage}{65mm} 
        \footnotesize \texttt{SELECT year FROM concert} \\
        \texttt{GROUP BY year} \\
        \texttt{ORDER BY count(*) DESC LIMIT 1}
        \end{minipage}
        \\
        \midrule
        \begin{tabular}{@{}l@{\;\;}}Ours \\ \cmark \end{tabular} & 
        \begin{minipage}{65mm} 
        \footnotesize
        \texttt{\#1 SELECT[concert.Year]}\\
        \texttt{\#2 PROJECT[concert, \#1]}\\
        \texttt{\#3 GROUP[count, \#2, \#1]}\\
        \texttt{\#4 SUPERLATIVE[max, \#1, \#3]}
        \end{minipage}
        \\
        \midrule
        Q: &  \begin{minipage}{65mm}
        \footnotesize What are the names of the stadiums without any concerts? %
        \end{minipage}\\
      \midrule      
        SQL: & 
        \begin{minipage}{65mm} 
        \footnotesize \texttt{SELECT name FROM stadium} \\
        \texttt{WHERE stadium\_id NOT IN} \\
        \texttt{(SELECT stadium\_id FROM concert)}
        \end{minipage}
        \\
        \midrule
        \begin{tabular}{@{}l@{\;\;}}Ours \\ \cmark \end{tabular} & 
        \begin{minipage}{65mm} 
        \footnotesize
        \texttt{\#1 SELECT[stadium]}\\
        \texttt{\#2 COMPARATIVE[\#1, \#1, concert]}\\
        \texttt{\#3 DISCARD[\#1, \#2]}\\
        \texttt{\#4 PROJECT[stadium.Name, \#3]}
        \end{minipage}
        \\
        \midrule
        Q: &  \begin{minipage}{65mm}
        \footnotesize What are the number of concerts that occurred in the stadium with the largest capacity? %
        \end{minipage}\\
       \midrule      
        SQL: & 
        \begin{minipage}{65mm} 
        \footnotesize
        \texttt{SELECT count(*) FROM concert}\\
        \texttt{WHERE stadium\_id =(}\\
        \texttt{SELECT stadium\_id FROM stadium} \\
        \texttt{ORDER BY capacity DESC LIMIT 1)}
        \end{minipage}
        \\
        \midrule
        \begin{tabular}{@{}l@{\;\;}}Ours \\ \xmark \end{tabular} & 
        \begin{minipage}{65mm} 
        \footnotesize
        \texttt{\#1 SELECT[stadium]}\\
        \texttt{\#2 PROJECT[stadium.Capacity, \#1]}\\
        \texttt{\#3 SUPERLATIVE[max, \#1, \#2]}\\
        \texttt{\#4 PROJECT[concert, \#3]}\\
        \texttt{\#5 AGGREGATE[sum, \#4]}
        \end{minipage}
        \\
        \midrule
        Q: &  \begin{minipage}{65mm}
        \footnotesize What is the average and maximum capacities for all stadiums? %
        \end{minipage}\\
      \midrule      
        SQL: & 
        \begin{minipage}{65mm} 
        \footnotesize \texttt{SELECT avg(capacity),}\\
        \texttt{max(capacity) FROM stadium}
        \end{minipage}
        \\
        \midrule
        \begin{tabular}{@{}l@{\;\;}}Ours \\ \xmark \end{tabular} & 
        \begin{minipage}{65mm} 
        \footnotesize
        \texttt{\#1 SELECT[stadium]}\\
        \texttt{\#2 PROJECT[stadium.Average, \#1]}\\
        \texttt{\#3 AGGREGATE[avg, \#2]}\\
        \texttt{\#4 AGGREGATE[max, \#2]}\\
        \texttt{\#5 UNION[\#3, \#4]}
        \end{minipage}
        \\
        \bottomrule
    \end{tabular}
   \caption{Qualitative results of our GraPPa-based model. \cmark and \xmark \space denote correct and incorrect execution results respectively. \label{tab:qdmr_results}}
\end{table}

\subsection{Additional training data from Break}
The Break dataset contains QDMR annotations for several question answering datasets, so we tried to enrich training on Spider with QDMRs from other parts of Break. %
Table~\ref{tab:full_break} shows the execution accuracy on our dev and test in these settings.
Adding training data for both versions of the model leads to performance improvement on the test set, but slightly decreases the dev set results.

When training with the data from other parts of Break, we
simply assume that the schema is empty and use all the textual QDMR arguments as values.
More careful exploration of additional QDMR data is left for future work.

\subsection{Ablation study}
Table~\ref{tab:ablation} presents results of ablations on
the development set.
First, note that disabling augmentations in both models decreases the execution accuracy.

Next, we tested different configurations of RAT-encoder: 
\begin{itemize}[noitemsep,topsep=0mm,leftmargin=*]
    \item without relations that come from the schema structure (e.g., the table~-- primary key and foreign key relations);
    \item with the small number of default relations: without distinguishing table, column or value, because these elements are considered as elements of one unified grounding type;
    \item the regular transformer instead of RAT.
\end{itemize}

The model without schema relations lost %
$8\%$ on dev, which shows that encoding schema with RAT-encoder is an important part of the model. This also limits the use of additional data from Break, where schemas do not exist. 
The variety of relations in RAT-encoder is also important, as RAT itself.
Our findings are consistent with the ablations of \citet{wang-etal-2020-rat}.

\begin{table}[t]
\centering
\normalsize
\begin{tabular}{@{}lccc@{}}
\toprule %
\textbf{Model} & \textbf{Pretrain}  & \textbf{Dev} \\ %
\midrule %
Base & BERT & 81.1 \\ %
- w/o augmentations & BERT & 76.0 \\ %
- w/o schema relations & BERT & 67.9 \\ %
- with default relations & BERT & 65.6 \\ %
- w/o relation-aware layers & BERT & 51.2 \\ %
\midrule %
Base & GraPPa  & 82.0  \\ %
- w/o augmentations & GraPPa  & 81.1 \\ %
\bottomrule %
\end{tabular}
\caption{Execution accuracy for our ablation study.}
\label{tab:ablation}
\end{table}

\section{Related Work} \label{sec:related_works}
\paragraph{Text-to-SQL.} 
The community has recently made significant progress and moved from fixed-schema datasets like ATIS or GeoQuery \citep{popescu2003,iyer-etal-2017-learning} to the WikiSQL or Overnight datasets with multiple single-table schemas~\citep{wang-etal-2015-building,zhongSeq2SQL2017} and then to the Spider dataset with multiple multi-table multi-domain schemas~\citep{yu-etal-2018-spider}.
Since the release of Spider, the accuracy has moved up from around 10\% to 70\%.

Most recent systems are structured as encoder-decoder networks.
Encoders typically consist of a module fine-tuned from a pretrained language model like BERT~\citep{devlin-etal-2019-BERT} and a module for incorporating the schema structure.
\citet{guo-etal-2019-towards,zhong-etal-2020-grounded,lin-etal-2020-bridging} represented schemas as token sequences, \citet{bogin-etal-2019-representing,bogin-etal-2019-global} used graph neural networks and \citet{wang-etal-2020-rat} used relation-aware transformer, RAT, to encode a graph constructed from an input schema.
In this paper, we use the RAT module to encode the schema but enlarge the encoded graph by adding value candidates as nodes.

Decoders are typically based on a grammar representing a subset of SQL and produce output tokens in the depth-first traversal order of an abstract syntax tree, AST, following~\citet{yin-neubig-2017-syntactic}.
A popular choice for such a grammar is to use SemQL of \citet{guo-etal-2019-towards} or to use a lighter grammar with more intensive consistency checks inside beam search like in BRIDGE~\citep{lin-etal-2020-bridging}.
Recently, \citet{rubin2020smbop} proposed a different approach to decoding based on bottom-up generating of sub-trees on top of the relational algebra of SQL.
In our paper, we follow the standard AST-based approach but for the grammar describing grounded QDMRs.
We also use some consistency checks and the decoding time to prevent some easily avoidable inconsistencies.
 
There is also a line of work on weakly-supervised learning of text-to-SQL semantic parsers, where SQL queries or logical forms for the training set are not available at all.
Some works \citep{min-etal-2019-discrete,wang-etal-2019-learning-semantic,agarwal2019,liang2018} reported results on the WikiSQL dataset, \citet{wang2021} worked on GeoQuery and Overnight datasets.
We are not aware of any works reporting weakly-supervised results on the multi-table Spider dataset.

\paragraph{Pretraining on text and tables.}
One possible direction inspired by the success of pretraining language models on large text corpora is to pretrain  model on data with semantically connected text and tables.
\citet[TaBERT]{yin-etal-2020-taBERT} and \citet[TaPas]{herzig-etal-2020-tapas} used text-table pairs extracted from sources like Wikipedia for pretraining.
\citet[GraPPa]{yu-etal-2021-GraPPa} used synthetic question-SQL pairs.
\citet[STRUG]{deng2021structure-grounded} used the table-to-text dataset of \citet[ToTTo]{parikh-etal-2020-totto}.
\citet[GAP]{shi-etal-2021-gap} used synthetic data generated by the models for SQL-to-text and table-to-text auxiliary tasks.
In this paper, we do not pretrain such models but experiment with GraPPa as the input encoder.

\paragraph{QDMR.}
Together with the Break dataset, \citet{wolfson-etal-2020-break} created a task of predicting QDMRs given questions in English.
As a baseline, they created a seq2seq model enhanced with a copy mechanism of \citet{gu-etal-2016-incorporating}.
Recently, \citet{hasson2021question} built a QDMR parser that is based on dependency graphs and uses RAT modules.
Differently from this line of work, we use a modified version of QDMRs, and our models never actually predict QDMR arguments as text but always directly their groundings.

\paragraph{SPARQL.}
SPARQL was used in several lines of work on semantic parsing for querying knowledge bases.
The SEMPRE system of~\citet{berant-etal-2013-semantic} relied on SPARQL to execute logical forms on the Freebase knowledge base.
\citet{yih-etal-2016-value}  and \citet{talmor-berant-2018-web} created the WebQuestions and  ComplexWebQuestions datasets, respecively, where annotations were provided in the form of SPARQL queries.
A series of challenges on Question Answering over Linked Data Challenge, QALD \citep{LOPEZ20133}, and the LC-QuAD datasets \citep{trivedi2017,dubey2019lc2} targeted the generation of SPARQL queries directly.
Our paper is different from these lines of work as we rely on supervision via QDMRs and not SPARQL directly.

There also exist several lines of works on converting queries from/to SPARQL, and the problems are difficult.
See, e.g., the works of \citet{Michel2019,abatal2019} and references therein.

\section{Conclusion} \label{sec:conclusion}
In this paper, we proposed a way to use the recent QDMR format~\citep{wolfson-etal-2020-break} as annotation for generating executable queries to databases given a question in a natural language.
Using QDMRs is beneficial because they can be collected through crowdsourcing potentially easier than correct database queries.
Our system consists of two main parts.
First, we have a learned text-to-QDMR translator that we built on top of the recent RAT-SQL system~\citep{wang-etal-2020-rat} and trained on an annotated with QDMRs part of the Spider dataset.
Second, we have a non-trainable QDMR-to-SPARQL translator, which generates queries executable on databases in the RDF format.
We evaluated our system on the Spider dataset and showed it to perform on par with the modern text-to-SQL methods (BRIDGE and SmBoP) trained with full supervision in the form of SQL queries.
We also showed that additional QDMR annotations for questions not aligned with any databases could further improve the performance.
The improvement shows great potential for future work.

\section*{Acknowledgements}
This research was supported in part through computational resources of HPC facilities at HSE University \citep{Kostenetskiy_2021}.

\bibliography{anthology,custom}
\bibliographystyle{acl_natbib}

\clearpage
\appendix

\vskip .375in
\begin{center}
    {\Large \bf Supplementary Material (Appendix) \par} {\Large \bf SPARQLing Database Queries from Intermediate Question Decompositions\par}
    \vspace*{10pt}
\end{center}

\section{QDMR-to-SPARQL translator}
\label{sec:qdmr2sparql}

Table~\ref{tab:qdmr_ops_full} contains the full list of QDMR operators used in our paper.

Algorithm~\ref{alg:qdmr2sparql} sketches the QDMR-to-SPARQL translator.
It is a recursive procedure that creates SPARQL queries for all QDMR LF steps.
At its core, it constructs one or several patterns for the step arguments and then connects them into another pattern in a way specific to the LF operator of the current step.

Importantly, the patterns for LF operators can be of  two types: inner (inline) and full.
An inner pattern represents the internal part of a query that needs to be placed inside the curly brackets \texttt{\{\dots$\!$\}}.
A full pattern corresponds to a full query that can be executed directly (starts with the \texttt{SELECT} keyword).
An inner pattern can be converted to full by using the \texttt{SELECT <output vars> WHERE \{<inner>\}} construction.
The full pattern can be converted to inner by creating a subquery via \texttt{\{<full>\}} (here, the output variables of \texttt{<full>} pattern become available in the scope where the subquery is created).

Different LF operators require and produce different patterns: inner of full.
Next, we specify a pattern for each LF operator.

The \texttt{SELECT} operator adds the grounded object to the context: a self link for a table, a link for a column, a link with a filtering condition for a value.

The \texttt{PROJECT} operator creates a context for the argument and does the same as \texttt{SELECT}.
To connect instances from different columns, we use the breadth-first search to find the shortest path in the undirected graph where all the columns of all tables represent nodes and edges appear between the primary key of each table and all other columns of the same table, and along with the foreign links.

The \texttt{COMPARATIVE} operator first creates an inner \texttt{<pattern>} for its arguments and then adds a filtering condition from the l.h.s.\ values \texttt{<filter\_var>}, the operation \texttt{<comparator>} and the r.h.s.\ value~\texttt{<value>}:
\begin{lstlisting}[columns=fullflexible,keepspaces=true,basicstyle=\ttfamily\footnotesize]
<pattern>
FILTER(<filter_var><comparator><value>).
\end{lstlisting}

\begin{algorithm}[t]
\caption{QDMR-to-SPARQL translator}
\label{alg:qdmr2sparql}
\begin{algorithmic}[1]
\Function{qdmr2sparql}{\;}
\State\Call{GetContext}{$[i_{\text{out}}]$, True, $\emptyset$}
\EndFunction
\Statex
\Function{GetContext}{indices, inline, $C$} 
    \If{C is empty} $C$ $\leftarrow$ \Call{InitC}{ } \EndIf
    \While{not too many tries}
    \For{$i$ \textbf{in} indices}
        \State $C$ $\leftarrow$ \Call{AddIndex}{$i$, inline, $C$}
    \EndFor
    \EndWhile
    \State \Return C
\EndFunction
\Statex
\Function{AddIndex}{$i$, inline, $C$}
    \State op $\leftarrow$ QDMR op at step \#$i$
    \State args $\leftarrow$ get indices of arguments of step \#$i$
    \State needs\_inl $\leftarrow$ [op needs inline args]
    \State makes\_inl $\leftarrow$ [op makes inline context]
    \State $C\leftarrow$ \Call{GetContext}{args, needs\_inl, $C$}
    \State $C \leftarrow$ fill the pattern of op
    \If{inline $\neq$ makes\_inl}
        \State $C \leftarrow$ convert $C$ to inline/full
    \EndIf
    \State \Return $C$
\EndFunction
\end{algorithmic}
\end{algorithm}

\begin{table*}[t!]
    \centering
    \footnotesize
    \begin{tabular}{@{}llll@{}}
        \toprule
         \textbf{Operator} & \textbf{Arguments} & \textbf{Type} & \textbf{Description} \\
        \midrule
         \multirow{2}{*}{\texttt{SELECT}} & \texttt{subject} & \texttt{text} & Select \texttt{subject}  \\
                                        & \texttt{distinct}  & \texttt{bool} & (possibly \texttt{distinct} values) \\
         \midrule
         \multirow{3}{*}{\texttt{PROJECT}} & \texttt{projection} & \texttt{text} & Select \texttt{projection}  \\
          & \texttt{subject} & \texttt{ref} & related to \texttt{subject} \\
          & \texttt{distinct} & \texttt{bool} & (possibly \texttt{distinct} values)\\
         \midrule
         \multirow{5}{*}{\texttt{COMPARATIVE}} & \texttt{subject} & \texttt{ref} & Select \texttt{subject} such that  \\
          & \texttt{attr} & \texttt{ref} & related \texttt{attr} compares \\
          & \texttt{comparator} & \texttt{choice} & (using $=$, $\neq$, $>$, $<$, $\geq$, $\leq$,  \texttt{like}) \\
          & \texttt{value} & \texttt{text}/\texttt{ref} & to \texttt{value} \\
          & \texttt{distinct} & \texttt{bool} & (possibly \texttt{distinct} values)\\
          \midrule
         \multirow{3}{*}{\texttt{SUPERLATIVE}} & \texttt{subject} & \texttt{ref} & Select \texttt{subject}  such that \\
          & \texttt{attr} & \texttt{ref} &  related \texttt{attr} has \\
          & \texttt{operator} & \texttt{choice} & $\max$/$\min$ values \\
         \midrule
         \multirow{2}{*}{\texttt{AGGREGATE}} & \texttt{subject} & \texttt{ref} & Compute $\max$/$\min$/\texttt{sum}/\texttt{count} \\
          & \texttt{operator} & \texttt{choice} & of \texttt{subject} \\
         \midrule
         \multirow{3}{*}{\texttt{GROUP}} & \texttt{subject} & \texttt{ref} & Group instances of \texttt{subject} \\
          & \texttt{attr} & \texttt{ref} & such that \texttt{attr} has same values \\
          & \texttt{operator} & \texttt{choice} & (aggregate with $\max$/$\min$/\texttt{sum}/\texttt{count}) \\
         \midrule
         \multirow{1}{*}{\texttt{UNION}} & \texttt{ref1}, \texttt{ref2}, etc. & \texttt{ref} & Get the union of \texttt{ref1}, \texttt{ref2}, etc. \\
         \midrule
         \multirow{2}{*}{\texttt{INTERSECT}} & \texttt{subject} & \texttt{ref} & Get instances of \texttt{subject} \\
          & \texttt{attr1}, \texttt{attr2} & \texttt{ref} & related to both \texttt{attr1} and \texttt{attr2} \\
         \midrule
         \multirow{2}{*}{\texttt{DISCARD}} & \texttt{subject} & \texttt{ref} & Get instances of \texttt{subject} \\
          & \texttt{minus} & \texttt{ref} & excluding instances of \texttt{minus} \\
         \midrule
         \multirow{3}{*}{\texttt{SORT}} & \texttt{subject} & \texttt{ref} & Order instances of \texttt{subject} \\
          & \texttt{attr} & \texttt{ref} & such that related \texttt{attr} \\
          & \texttt{direction} & \texttt{choice} & is ordered in \texttt{asc}/\texttt{desc} direction \\
         \bottomrule
    \end{tabular}
    \caption{QDMR operators, their arguments, types of the arguments. The full version of Table~\ref{tab:qdmr_ops}.}
    \label{tab:qdmr_ops_full}
\end{table*}

The \texttt{AGGREGATE} operator computes the aggregator \texttt{<agg\_op>} from a set of values.
This operator takes the inner pattern \texttt{<pattern>} as input (with \texttt{<var>} correspondings to the set of values to aggregate) and produces the full query with the output variable \texttt{<output\_var>} as the output:
\begin{lstlisting}[columns=fullflexible,keepspaces=true,basicstyle=\ttfamily\footnotesize]
SELECT (<agg_op>(<var>) as <output_var>)
WHERE { <pattern> }
\end{lstlisting}

The \texttt{SUPERLATIVE} operator filters the instances such that some related attribute has the min/max value.
The operator first computes the min/max value with a built-in \texttt{AGGREGATE} operator then filters (similar to \texttt{COMPARATIVE}) the patterns based on the computed value:
\begin{lstlisting}[columns=fullflexible,keepspaces=true,basicstyle=\ttfamily\footnotesize]
{SELECT (<agg_op>(<var>) AS <minmax_var>)
  WHERE { <pattern_inner> } } 
<pattern_outer>
FILTER(<query_outer_var>=<minmax_var>).
\end{lstlisting}
The \texttt{SUPERLATIVE} operator requires two inner patterns as input \texttt{<pattern\_inner>}, \texttt{<pattern\_outer>} and makes an inner pattern as the output.
        
The \texttt{GROUP} operator groups the values \texttt{<var>} by the equal values of the related attribute \texttt{<index\_var>}:
\begin{lstlisting}[columns=fullflexible,keepspaces=true,basicstyle=\ttfamily\footnotesize]
SELECT (<agg_op>(<var>) AS <output_var>)
WHERE { <pattern> }
GROUP BY <index_var>
\end{lstlisting}
The aggregation is done with the operator \texttt{<agg\_op>}.
The input pattern \texttt{<pattern>} is inner, and the output is the full pattern with the output variable \texttt{<output\_var>}.

The \texttt{UNION} operator can actually correspond to several operators: horizontal union, vertical union, union of aggregators, union after group.
By horizontal union, we mean the union of two or more related variables from the same pattern.
These variables have to correspond to different database columns.
By vertical union, we mean the union of two or more variables corresponding to the same column but coming from different patterns.
This case is implemented with the \texttt{UNION} keyword from SPARQL using the following construction:
\begin{lstlisting}[columns=fullflexible,keepspaces=true,basicstyle=\ttfamily\footnotesize]
{ <pattern1> }
UNION
{ <pattern2> }
\end{lstlisting}
The union-after-group case is a special but common situation when arguments contain the result of the \texttt{GROUP} operator and the index variable of the same operator.
We implement this case similar to the pattern of the \texttt{GROUP} operator but with several variables in the output.
The union of aggregators is another common special case when the arguments of the \texttt{UNION} contain several aggregators from the same pattern.
We simply output these several aggregators by concatenating them after the SPARQL \texttt{SELECT} keyword.

The \texttt{INTERSECT} operator effectively consists in sequentially applying two \texttt{COMPARATIVE} operators that do not have explicit comparisons as arguments.

The \texttt{DISCARD} operator is based of the pattern very similar to the vertical union:
\begin{lstlisting}[columns=fullflexible,keepspaces=true,basicstyle=\ttfamily\footnotesize]
{ <pattern1> }
MINUS
{ <pattern2> }
\end{lstlisting}

The \texttt{SORT} operator consists in adding the \texttt{ORDER BY} \  keyword at the end of the full pattern:
\begin{lstlisting}[columns=fullflexible,keepspaces=true,basicstyle=\ttfamily\footnotesize]
SELECT <output_vars>
WHERE { <pattern> }
ORDER BY ASC/DESC(<index_var>)
\end{lstlisting}
\newpage

\section{Implementation details}
\label{sec:implementation}
We implemented our model on the top of the RAT-SQL code\footnote{\url{https://github.com/microsoft/rat-sql}} built with Pytorch \citep{NEURIPS2019_9015}.
We use pretrained BERT and GraPPa from the Transformers library   \citep{wolf-etal-2020-transformers}.
To support SPARQL queries and RDF databases, we used two libraries: RDFLib\footnote{\url{https://github.com/RDFLib/rdflib}} and the open-source version of the Virtuoso system.\footnote{\url{https://github.com/openlink/virtuoso-opensource}}
RDFLib was much easier to install (a python package), but Virtuoso allowed to run SPARQL queries on pre-loaded databases much faster.

To choose relevant values from a database, we tokenized question and all unique database values using the Stanford CoreNLP library \citep{manning-etal-2014-stanford}, filtered tokens using NLTK\footnote{\url{https://www.nltk.org/index.html}} English stopwords, and then picked top-25 values with higher similarity scores calculated as follows:
\begin{itemize}
    \item for a numeric value, we gave the maximum score if it exactly matched with some question token, otherwise, we gave the minimum score;
    \item for other tokens, we gave the maximum score if the value and question stems were the same (we used the Porter and Snowball stemmers from NLTK), otherwise, we calculated similarity score based on the longest continuous matching subsequence (we used Python SequenceMatcher class).
\end{itemize}

For the neural network architecture and training, we used the same hyperparameters as RAT-SQL \citep{wang-etal-2020-rat}: 8 RAT layers, each with 8 heads and the hidden dimension of 256, 1024 and 512 in self-attention, position-wise feed-forward network and decoder LSTM, respectively.
We trained the model with the Adam optimizer \citep{kingma2014adam} and polynomial decay scheduler used by \citet{wang-etal-2020-rat}.
The batch size was 24, the overall number of iterations was 81000 for all models.

The training time on 4 NVIDIA V100 GPUs was approximately 24 hours.

\vfill %

\section{QDMR grammar}
\label{sec:grammar}
\small
\begin{tabular}{@{}l@{\:}c@{\;\;}l@{}}  
\texttt{root} & $\longrightarrow$ & \texttt{step} \\
\texttt{step} & $\longrightarrow$ & \texttt{select, project, sort,} \\
& & \texttt{group, aggregate,} \\
& & \texttt{comparative, superlative,}\\
& & \texttt{intersection, discard,} \\
& & \texttt{union, final} \\
\texttt{select} & $\longrightarrow$ & \texttt{distinct, grounding, step} \\ 
\texttt{project} & $\longrightarrow$ & \texttt{distinct, project\_1arg}\\
& & \texttt{ref, step} \\ 
\texttt{comparative} & $\longrightarrow$ & \texttt{distinct, ref, ref,} \\
& & \texttt{comp\_3arg, step} \\ 
\texttt{superlative} & $\longrightarrow$ & \texttt{superlative\_op}\\
& & \texttt{ref, ref, step} \\ 
\texttt{group} & $\longrightarrow$ & \texttt{agg\_type, ref, ref, step} \\ 
\texttt{aggregate} & $\longrightarrow$ & \texttt{agg\_type, ref, step} \\ 
\texttt{intersection} & $\longrightarrow$ & \texttt{ref, ref, step} \\ 
\texttt{discard} & $\longrightarrow$ & \texttt{ref, ref, step} \\
\texttt{union} & $\longrightarrow$ & \texttt{ref, ref, step} \\
\texttt{sort} & $\longrightarrow$ & \texttt{ref, ref, order, step} \\
\texttt{project\_1arg} & $\longrightarrow$ & \texttt{grounding} | \texttt{none} \\
\texttt{comp\_3arg} & $\longrightarrow$ & \texttt{comp\_op\_type}, \\ & & \texttt{column\_type, comp\_val} \\
\texttt{comp\_op\_type} & $\longrightarrow$ & \texttt{comparative\_op} | \texttt{no\_op} \\
\texttt{column\_type} & $\longrightarrow$ & \texttt{grounding} | \texttt{no\_column} \\
\texttt{comp\_val} & $\longrightarrow$ & \texttt{grounding} | \texttt{ref} \\
\texttt{comparative\_op} & $\longrightarrow$ & $\ne | > | <|  \ge | \le | $\texttt{like} \\
\texttt{superlative\_op} & $\longrightarrow$ & $\min | \max$ \\
\texttt{order} & $\longrightarrow$ & \texttt{asc} | \texttt{desc} \\
\texttt{agg\_type}& $\longrightarrow$ & \texttt{aggregate\_op} | \texttt{grounding}  \\
\texttt{aggregate\_op} & $\longrightarrow$ & \texttt{Count} | \texttt{Sum} | \texttt{Avg} | \\ & & \texttt{Min} | \texttt{Max}  \\
\end{tabular}
\normalsize

\section{Constraints in the decoding process}
\label{sec:decoding_rules}
The decoding process at the inference stage is sequential, and at each step, there is a set of eligible choices. These sets are always non-empty and are formed using the following constraints:
\begin{itemize}
\item The eligible choices of grounding as aggregate type ({\small\texttt{agg\_type} $\longrightarrow$ \texttt{grounding}}) columns;
\item The eligible choices of grounding as the column type in comparative ({\small\texttt{column\_type} $\longrightarrow$ \texttt{grounding}}) are columns with the types from the set of input value types;
\item After the model chooses a column in comparative, the eligible choices of grounding as comparative value ({\small\texttt{comp\_val} $\longrightarrow$ \texttt{grounding}})  are the values from this column or with the same type but not from the database;
\item After the model chose to skip column ({\small\texttt{no\_column}}) in comparative, the eligible choices of grounding as comparative value ({\small\texttt{comp\_val} $\longrightarrow$ \texttt{grounding}}) are the values not from the database.
\end{itemize}

\end{document}